\documentclass[10pt,journal,compsoc]{IEEEtran}

\ifCLASSOPTIONcompsoc
  \usepackage[nocompress]{cite}
\else
  \usepackage{cite}
\fi

\ifCLASSINFOpdf
\else
\fi

\usepackage{graphicx}
\usepackage{comment}
\usepackage{subfigure}
\usepackage{wrapfig}
\usepackage{multirow}
\usepackage{arydshln}
\usepackage{amsmath,amssymb} 
\usepackage{color}
\usepackage{hyperref}
\usepackage{url}

\def\ie{{\it i.e.}}
\def\eg{{\it e.g.}}
\def\et{{\it et al.}}

\def\f{\mathbf{f}}
\def\t{\mathbf{t}}

\def\M{{\mathbf M}}

\def\V{{\mathbf V}}
\def\X{{\mathbf X}}
\def\Y{{\mathbf Y}}

\begin{document}

\title{Self-Supervised Multi-View Learning via Auto-Encoding 3D Transformations}
\author{Xiang~Gao,~\IEEEmembership{Student Member,~IEEE,}
        Wei~Hu,~\IEEEmembership{Member,~IEEE,}
        and~Guo-Jun~Qi,~\IEEEmembership{Senior~Member,~IEEE}
\IEEEcompsocitemizethanks{
\IEEEcompsocthanksitem X. Gao and W. Hu are with Wangxuan Institute of Computer Technology, Peking University, No. 128 Zhongguancun North Street, Beijing, China.\protect\\
E-mail: \{gyshgx868, forhuwei\}@pku.edu.cn
\IEEEcompsocthanksitem G.-J. Qi is with the Futurewei Seattle Cloud Lab, Seattle, WA.\protect\\
E-mail: guojunq@gmail.com
\IEEEcompsocthanksitem The corresponding author is W. Hu (forhuwei@pku.edu.cn).
}
}

\markboth{Gao \MakeLowercase{\textit{et al.}}: Self-Supervised Multi-View Learning via Auto-Encoding 3D Transformations}{Journal of \LaTeX\ Class Files,~Vol.~14, No.~8, August~2015}%

%

\IEEEtitleabstractindextext{%
\begin{abstract}
3D object representation learning is a fundamental challenge in computer vision to infer about the 3D world.
Recent advances in deep learning have shown their efficiency in 3D object recognition, among which view-based methods have performed best so far.
However, feature learning of multiple views in existing methods is mostly performed in a supervised fashion, which often requires a large amount of data labels with high costs.
In contrast, self-supervised learning aims to learn multi-view feature representations without involving labeled data.
To this end, we propose a novel self-supervised paradigm to learn Multi-View Transformation Equivariant Representations (MV-TER), exploring the equivariant transformations of a 3D object and its projected multiple views. 
Specifically, we perform a 3D transformation on a 3D object, and obtain multiple views before and after the transformation via projection.  
Then, we self-train a representation to capture the intrinsic 3D object representation by decoding 3D transformation parameters from the fused feature representations of multiple views before and after the transformation.
Experimental results demonstrate that the proposed MV-TER significantly outperforms the state-of-the-art view-based approaches in 3D object classification and retrieval tasks, and show the generalization to real-world datasets.
\end{abstract}

}

\maketitle

\IEEEdisplaynontitleabstractindextext
\IEEEpeerreviewmaketitle

\IEEEraisesectionheading{\section{Introduction}
\label{sec:intro}}

3D object representation has become increasingly prominent for a wide range of applications, such as 3D object recognition and retrieval \cite{maturana2015voxnet,qi2016volumetric,brock2016generative,qi2017pointnet,qi2017pointnet++,klokov2017escape,su2015multi,feng2018gvcnn,yu2018multi,yang2019learning}.
Recent advances in Convolutional Neural Network (CNN) based methods have shown their success in 3D object recognition and retrieval \cite{su2015multi,feng2018gvcnn,yu2018multi,yang2019learning}.
One important family of methods are view-based methods, which project a 3D object into multiple views and learn compact 3D representation by fusing the feature maps of these views for downstream tasks.
Feature learning of multiple views in existing approaches is mostly trained in a supervised fashion, hinging on a large amount of data labels that prevents the wide applicability. 
Hence, self-supervised learning is in demand to alleviate the dependencies on labels by exploring unlabeled data for the training of multi-view feature representations in an unsupervised or (semi-)supervised fashion.

Many attempts have been made to explore self-supervisory signals at various levels of visual structures for representation learning.
The self-supervised learning framework requires only unlabeled data in order to formulate a \textit{pretext} learning task \cite{kolesnikov2019revisiting}, where a target objective can be computed without any supervision.
These pretext tasks can be summarized into four categories \cite{jing2019self}: generation-based \cite{zhang2016colorful,pathak2016context,srivastava2015unsupervised}, context-based, free semantic label-based \cite{faktor2014video,stretcu2015multiple,ren2018cross}, and cross modal-based \cite{sayed2018cross,korbar2018cooperative}.
Among them, context-based pretext tasks include representation learning from image transformations, which is well connected with transformation equivariant representations as they transform equivalently as the transformed images.    

Transformation Equivariant Representation learning assumes that representations equivarying to transformations are able to encode the intrinsic structures of data such that the transformations can be reconstructed from the representations before and after transformations \cite{qi2019learning}.
Learning transformation equivariant representations has been advocated in Hinton's seminal work on learning transformation capsules \cite{hinton2011transforming}.
Following this, a variety of approaches have been proposed to learn transformation equivariant representations \cite{kivinen2011transformation,sohn2012learning,schmidt2012learning,skibbe2013spherical,lenc2015understanding,gens2014deep,dieleman2015rotation,dieleman2016exploiting,zhang2019aet,qi2019avt,gao2020graphter,wang2020transformation}.
Specifically, Zhang \et \cite{zhang2019aet} propose to learn unsupervised representations of single images via Auto-Encoding Transformations (AET) by decoding transformation parameters from the learned representations of both the original and transformed images.
Further, Gao \et \cite{gao2020graphter} extend transformation equivariant representations to graph data that are irregularly structured (\eg, 3D point clouds), and formalize graph transformation equivariant representation learning by auto-encoding node-wise transformations in an unsupervised manner.
Nevertheless, these works focus on transformation equivariant representation learning of a single modality, such as 2D images or 3D point clouds.  

In this paper, we propose to learn Multi-View Transformation Equivariant Representations (MV-TER) by decoding the 3D transformations from multiple 2D views. 
This is inspired by the equivariant transformations of a 3D object and its projected multiple 2D views. 
That is, when we perform 3D transformations on a 3D object, the 2D views projected from the 3D object via fixed viewpoints will transform equivariantly. 
In contrast to previous works where 2D/3D transformations are decoded from the original single image/point cloud and transformed counterparts, we exploit the equivariant transformations of a 3D object and the projected 2D views. 
We propose to decode 3D transformations from multiple views of a 3D object before and after transformation, which is taken as self-supervisory regularization to enforce the learning of intrinsic 3D representation.  
By estimating 3D transformations from the fused feature representations of multiple original views and those of the equivariantly transformed counterparts from the same viewpoints, we enable the accurate learning of 3D object representation even with limited amount of labels.  

Specifically, we first perform 3D transformation on a 3D object (\eg, point clouds, meshes), and render the original and transformed 3D objects into multiple 2D views with fixed camera setup.
Then, we feed these views into a representation learning module to infer representations of the multiple views before and after transformation respectively. 
A decoder is set up to predict the applied 3D transformation from the fused representations of multiple views before and after transformation.
We formulate multi-view transformation equivariant representation learning as a regularizer along with the loss of a specific task (\eg, classification) to train the entire network end-to-end.
Experimental results demonstrate that the proposed method significantly outperforms the state-of-the-art view-based models in 3D object classification and retrieval tasks.

The proposed method distinguishes from AET \cite{zhang2019aet} in two aspects: 1) AET aims to learn equivariant representations of {\it single images} by estimating the applied {\it 2D transformations}. In contrast, we focus on representations equivarying to projected {\it 3D transformations} onto {\it multiple 2D views} by estimating the applied 3D transformation; 2) While the projection operator varies to different viewpoints, the acquired representations of 2D views share the same 3D transformation. In other words, the 3D transformation {\it links} the learned features of all the views. By decoding the 3D transformation from the feature representations of multiple 2D views, the model learns 3D information that reveals the intrinsic structure of the 3D object.

Our main contributions are summarized as follows.
\begin{itemize}
    \item We propose Multi-View Transformation Equivariant Representations (MV-TER) to learn 3D object representations from multiple 2D views that transform equivariantly with the 3D transformation in a self-supervised fashion.
    \item We formalize the MV-TER as a self-supervisory regularizer to learn the 3D object representations by decoding 3D transformation from fused features of projected multiple views before and after the 3D transformation of the object.  
    \item Experiments demonstrate the proposed method outperforms the state-of-the-art view-based methods in 3D object classification and retrieval tasks in a self-supervised fashion.
\end{itemize}

The remainder of this paper is organized as follows. We first review related works in Section~\ref{sec:related_works}. Then we formalize our model and provide analysis in Section~\ref{sec:formulation}, and discuss the proposed algorithm in Section~\ref{sec:method}. Finally, experimental results and conclusions are presented in Section~\ref{sec:experiments} and Section~\ref{sec:conclusion}, respectively.

\section{Related Works}
\label{sec:related_works}

In this section, we review previous works on self-supervised representation learning, transformation equivariant representations, as well as multi-view based neural networks.

\subsection{Self-Supervised Representation Learning}

Many self-supervised learning approaches have been proposed to train deep neural networks using self-supervised signals, which can be derived from data themselves without being manually labeled.
These self-supervised frameworks only require unlabeled data to formulate a pretext learning task \cite{kolesnikov2019revisiting}.
These pretext tasks can be summarized into four categories \cite{jing2019self}: generation-based, context-based, free semantic label-based, and cross modal-based.
Generation-based methods learn visual features by solving pretext tasks that involve image generation \cite{zhang2016colorful,pathak2016context} or video prediction \cite{srivastava2015unsupervised}.
Context-based pretext tasks mainly employ the context features of images or videos to train neural networks, such as context similarity \cite{noroozi2018boosting,caron2018deep}, spatial structure \cite{noroozi2016unsupervised,kim2018learning,doersch2015unsupervised}, or temporal structure \cite{misra2016shuffle,wei2018learning}.
Free semantic label-based methods train neural networks with automatically generated semantic labels.
The labels are generated by traditional hardcode algorithms \cite{faktor2014video,stretcu2015multiple} or by game engines \cite{ren2018cross}.
Cross modal-based methods aim to train convolutional neural networks to verify whether two different channels of input data are corresponding to each other \cite{sayed2018cross,korbar2018cooperative}.

Contrastive learning instantiates a family of self-supervised methods \cite{hjelm2019learning,velickovic2019deep,bachman2019learning,chen2020simple,he2020momentum,chen2020improved}, which maximizes the agreements between the augmented views of the same image in an embedding feature space, while avoiding the mode collapse of the embedded features by maximizing the disagreements between negative examples constructed from different images. Recently, an adversarial approach \cite{hu2020adco} is presented to demonstrate the negative pairs of examples can be directly trained end-to-end together with the backbone network so that the contrastive model can be learned more efficiently as a whole. 
In summary, these approaches employ self-supervised signals to train deep neural networks instead of manually labeled data.

\subsection{Transformation Equivariant Representations}

Many approaches have been proposed to learn equivariant representations, including transforming auto-encoders \cite{hinton2011transforming}, equivariant Boltzmann machines \cite{kivinen2011transformation,sohn2012learning}, equivariant descriptors \cite{schmidt2012learning}, and equivariant filtering \cite{skibbe2013spherical}.
Lenc \textit{et al.} \cite{lenc2015understanding} prove that the AlexNet \cite{krizhevsky2012imagenet} trained on ImageNet learns representations that are equivariant to flip, scaling and rotation transformations.
Gens \textit{et al.} \cite{gens2014deep} propose an approximately equivariant convolutional architecture, which utilizes sparse and high-dimensional feature maps to deal with groups of transformations.
Dieleman \textit{et al.} \cite{dieleman2015rotation} show that rotation symmetry can be exploited in convolutional networks for effectively learning an equivariant representation.
Dieleman \textit{et al.} \cite{dieleman2016exploiting} extend this work to evaluate on other computer vision tasks that have cyclic symmetry.
Cohen \textit{et al.} \cite{cohen2016group} propose group equivariant convolutions that have been developed to equivary to more types of transformations. The idea of group equivariance has also been introduced to the capsule nets \cite{lenssen2018group} by ensuring the equivariance of output pose vectors to a group of transformations.

To generalize to generic transformations, Zhang \textit{et al.} \cite{zhang2019aet} propose to learn unsupervised feature representations via Auto-Encoding Transformations (AET) by estimating transformations from the learned feature representations of both the original and transformed images.
Qi \textit{et al.} \cite{qi2019avt} extend AET by introducing a variational transformation decoder, where the AET model is trained from an information-theoretic perspective by maximizing the lower bound of mutual information.
Gao \textit{et al.} \cite{gao2020graphter} extend AET to graph data that are irregularly structured, and formalize graph transformation equivariant representation learning by auto-encoding node-wise transformations.
Wang \textit{et al.} \cite{wang2020transformation} extend the AET to Generative Adversarial Networks (GANs) for unsupervised image synthesis and representation learning.

\subsection{Multi-View Learning}

Recently, many view-based approaches have been proposed for 3D object learning.
These methods project 3D objects (\textit{e.g.}, point clouds, meshes) into multiple views and extract view-wise features receptively via CNNs, and then fuse these features as the descriptor of 3D objects.
Su \textit{et al.} \cite{su2015multi} first propose a multi-view convolutional neural network (MVCNN) to learn a compact descriptor of an object from multiple views, which fuses view-wise features via a max pooling layer.
Qi \textit{et al.} \cite{qi2016volumetric} introduce a new multi-resolution component into MVCNNs, and improve the classification performance.
However, max pooling only retains the maximum elements from views, which leads to information loss.
In order to address this problem, many subsequent works have been proposed to fuse multiple view-wise features into an informative descriptor for 3D objects.
Feng \textit{et al.} \cite{feng2018gvcnn} propose a group-view convolutional neural network (GVCNN) framework, which produces a compact descriptor from multiple views using a grouping strategy.
Yu \textit{et al.} \cite{yu2018multi} propose a multi-view harmonized bilinear network (MHBN), which learns 3D object representation by aggregating local convolutional features through the proposed bilinear pooling.

To take advantage of the spatial relationship among views, Han \textit{et al.} \cite{han2018seqviews2seqlabels} and Han \textit{et al.} \cite{han20193d2seqviews} propose to aggregate the global features of sequential views via attention-based RNN and CNN, respectively.
Kanezaki \textit{et al.} \cite{kanezaki2018rotationnet} propose to learn global features by treating pose labels as latent variables which are optimized to self-align in an unsupervised manner.
Yang \textit{et al.} \cite{yang2019learning} propose a relation network to connect corresponding regions from different viewpoints, and reinforce the information of individual views.
Jiang \textit{et al.} \cite{jiang2019mlvcnn} propose a Multi-Loop-View Convolutional Neural Network (MLVCNN) for 3D object retrieval by introducing a novel loop normalization to generate loop-level features.
Wei \textit{et al.} \cite{wei2020viewgcn} design a view-based graph convolutional network (GCN) framework to aggregate multi-view features by investigating relations of views.

\section{MV-TER: The Formulation}
\label{sec:formulation}

In this section, we first define multi-view equivariant transformation in Section~\ref{subsec:equivariant}.
Then we formulate the MV-TER model and introduce two transformation decoding schemes in Section~\ref{subsec:formulation} and Section~\ref{subsec:decoding}, respectively.
Further, some analysis and discussion of the proposed MV-TER are provided in Section \ref{subsec:discuz}.

\subsection{Multi-View Equivariant Transformation}
\label{subsec:equivariant}

2D views are projections of a 3D object from various viewpoints, which transform in an equivariant manner as the 3D object transforms.  
Formally, given a 3D object $\M \in \mathbb{R}^{n \times 3}$ consisting of $n$ points and a 3D transformation distribution $\mathcal{T}$, we sample a transformation $\t \sim \mathcal{T}$ and apply it to $\M$:
\begin{equation}
    \widetilde{\M} = \t(\M).
\end{equation}
We project $\M$ onto 2D views from $m$ viewpoints, denoted as $\mathcal{V}=\{\V_1,...,\V_m\}$,  \ie, 
\begin{equation}
    \V_i = p_i(\M),
    \label{eq:view_before_transformation}
\end{equation}
where $p_i: \mathbb{R}^3 \mapsto \mathbb{R}^2$ is a projection function for the $i$th view.
Subsequent to the transformation on $\M$, the $m$ views transform {\it equivariantly}, leading to $\widetilde{\mathcal{V}}=\{\widetilde{\V}_1,...,\widetilde{\V}_m\}$. 
We have 
\begin{equation}
    \widetilde{\V}_i = p_i\left(\widetilde{\M}\right) = p_i\left(\t(\M)\right)=\f_{i,t}\left(\V_i\right), i = 1,...,m,
    \label{eq:view_after_transformation}
\end{equation}
where $\f_{i,t}$'s are 2D transformations that are equivariant under the same 3D transformation $\t$. 
Though $\V_i$ and $\widetilde{\V}_i$ are projected along the same viewpoint $i$ (\ie, the same camera setup), they are projections of the original 3D object and its transformed counterpart, thus demonstrating different perspectives of the same 3D object.  
Our goal is to learn the representations of 3D objects from their multiple 2D views by estimating the 3D transformation $\t$ from sampled multiple views before and after the transformation, \ie, $\mathcal{V}$ and $\widetilde{\mathcal{V}}$.


\subsection{The Formulation}
\label{subsec:formulation}



Considering the $i$th 2D views $\{\V_i,\widetilde{\V}_i\}$ before and after a 3D transformation $\mathbf t$ applied to the corresponding 3D object, 
a function $E(\cdot)$ is \textit{transformation equivariant} if it satisfies 
\begin{equation}
    E(\widetilde{\V}_i) = E(\f_{i,t}(\V_i)) = \rho(\t)E(\V_i),
    \label{eq:ter}
\end{equation}
where $\rho(\t)$ is a homomorphism of transformation $\t$ in the representation space.

We aim to train a shared representation module $E(\cdot)$ that learns equivariant representations of multiple views. 
In the setting of self-supervised learning, we formulate MV-TER as a regularizer along with the (semi-)supervised loss of a specific task to train the entire network. 
Given a neural network with learnable parameters $\Theta$, the network is trained end-to-end by minimizing the weighted sum of two loss functions: 
1) the loss of a specific task $\ell_{\text{task}}$ (\eg, a cross-entropy loss in 3D object classification); and 
2) the MV-TER loss that is the expectation of estimation error $\ell_{\M}(\t, \hat{\t})$ over each sample $\M$ given a distribution of 3D objects $\mathcal M$ and each transformation $\t \sim \mathcal T$: 
\begin{equation}
    \min_{\Theta} \;
    \ell_{\text{task}} + \lambda \underset{\t \sim \mathcal{T}}{\mathbb E}~\underset{\M \sim \mathcal M}{\mathbb E}
     ~ \ell_{\M}(\t, \hat{\t}).
    \label{eq:loss}
\end{equation}
$\ell_{\M}(\t, \hat{\t})$ is the mean squared error (MSE) between the estimated transformation $\hat \t$ and the ground truth $\t$. 
$\lambda$ is a weighting parameter to strike a balance between the loss of a specific task and the MV-TER loss.  
Here, the loss $\ell_{\text{task}}$ can be taken over all the data labels (fully-supervised) or partial labels (semi-supervised).  
In \eqref{eq:loss}, $\hat \t$ is decoded as a function of $\mathcal V$ and $\widetilde{\mathcal V}$ in multiple views as defined in \eqref{eq:view_before_transformation} and \eqref{eq:view_after_transformation}, and we will present two schemes to decode $\hat \t$ in the next subsection. 

\begin{figure*}[t]
    \centering
    \includegraphics[width=0.95\textwidth]{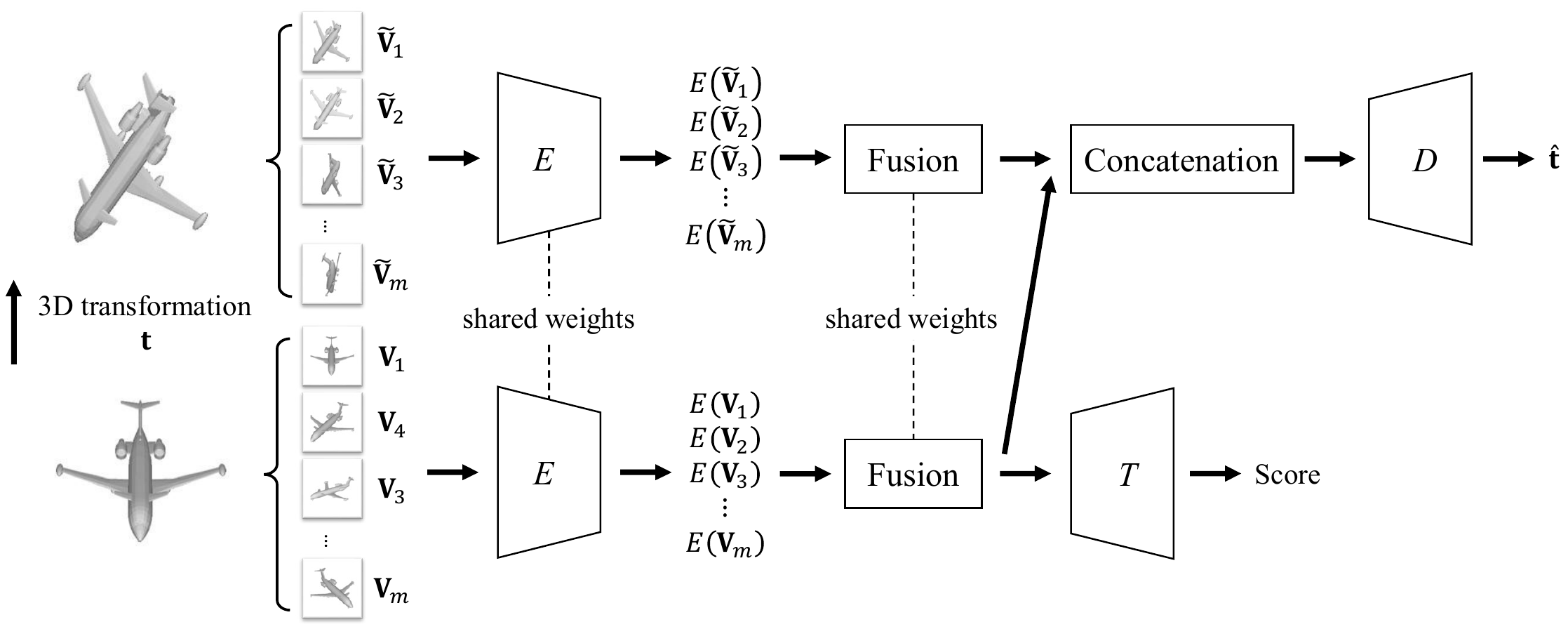}
    \caption{\textbf{The architecture of the proposed MV-TER in the fusion decoding scheme.} $E$ and $D$ represent the feature representation module and transformation decoding module respectively, and $T$ is a specific task, \textit{e.g.}, 3D object classification or retrieval.}
    \label{fig:framework_fusion}
\end{figure*}


\subsection{Two Transformation Decoding Schemes}
\label{subsec:decoding}

We propose two schemes to decode the transformation $\t$ in \eqref{eq:loss} from the feature representations of multiple views $E(\V_i)$ and $E(\widetilde \V_i), i=1,...,m$. 

\subsubsection{Fusion Scheme}
The first scheme is to decode from fused representations of multiple views (before and after transformations). 
Suppose the neural network extracts features of $\V_i$ and $\widetilde{\V}_i$ from a representation learning module $E(\cdot)$, and estimates the 3D transformation from both features via a transformation decoding module $D(\cdot)$, then we have 
\begin{equation}
  \hat{\t}=D\left[F\left(E(\V_1),...,E(\V_m)\right), F\left(E(\widetilde{\V}_1),...,E(\widetilde{\V}_m)\right)\right], 
  \label{eq:decode_fusion}
\end{equation}
where $F(\cdot)$ is a function of feature fusion. The fused feature of multiple views essentially represents the 3D object representation.

\subsubsection{Average Scheme}
In the second decoding scheme, we estimate the transformation $\hat{\t}$ from each view before and after transformation, and then take the average of the estimates. The idea is each view captures the projected 3D structures under a transformation. This essentially models a 3D object from different perspectives from which the underlying 3D transformation can be revealed. By averaging the estimated transformations across multiple views, a reasonable estimation of the 3D transformation can be made. 

This actually pushes the model to learn a good 3D representation from individual 2D views and leads to an estimation $\hat{\t}_i$ from the $i$th view:
\begin{equation}
    \hat{\t}_i=D\left[E(\V_i), E(\widetilde{\V}_i)\right], i=1,...,m.
    \label{eq:t_i}
\end{equation}
The final decoded 3D transformation is taken as the expectation of $\hat{\t}_i$'s:
\begin{equation}
  \hat{\t} = \frac{1}{m}\sum_{i=1}^m \hat{\t}_i.
  \label{eq:final}
\end{equation}
Hence, we update the parameters $\Theta$ in the representation learning module $E(\cdot)$ and the transformation decoding module $D(\cdot)$ iteratively by backward propagation of the regularized loss in \eqref{eq:loss}.  

Interestingly, the second decoding scheme can reach an even better performance than the first scheme in experiments.  This should not be surprising since our ultimate goal is to enable multi-view learning by fusing the representations of individual 2D views to reveal the target 3D objects.  The second scheme follows this motivation by pushing each view to encode as much information as possible about the 3D transformation, as implied by the multi-view learning.

\subsection{Discussion}
\label{subsec:discuz}

We provide some analysis and discussion on how the MV-TER loss helps to learn transformation equivariant representations, the difference between data augmentation and MV-TER, as well as the connection between multi-view transformation equivariance and 3D transformation equivariance.

\subsubsection{The MV-TER Loss} 

We minimize the MV-TER loss to enforce the property of the transformation equivariance as in \eqref{eq:ter}.
Specifically, we implement the definition by decoding the 3D transformation parameters $\t$ from the feature representations of multiple views before and after the transformation, \ie, $E(\V_i)$ and $E(\widetilde{\V}_i)$.
The proposed MV-TER loss measures the estimation accuracy of the 3D transformation, thus minimizing the MV-TER loss leads to the learning of transformation equivariant representations.

\subsubsection{Difference between Data Augmentation and MV-TER}

Compared with data augmentation where the transformed samples are used as additional training samples, the proposed MV-TER differs in the following three aspects:
\begin{enumerate}
  \item Data augmentation needs to mark different augmented versions of a sample with the same label, while the proposed self-supervised MV-TER does not require the label information of transformed samples.
  \item Data augmentation needs to ensure the ``class invariance" of samples, which limits the types of data augmentation (\eg, rotation-based data augmentation cannot be used in handwriting digit recognition, because ``6" would be rotated to ``9"). In comparison, the proposed MV-TER aims at modeling transformation equivariance, which admits a larger family of 3D transformations used in the MV-TER loss that measures the estimation accuracy of the applied transformations.
  \item The property of transformation equivariant representations is not guaranteed  by simply adopting data augmentation, as the transformation equivariance and class invariance are two different concepts in unsupervised and supervised settings respectively.
\end{enumerate}

\subsubsection{Connection to 3D Transformation Equivariance}

Another perspective of the transformation decoding schemes takes inspiration from {\it 3D Transformation Equivariance}. 
Taking the Fusion Scheme as an example, as the 3D object $\M$ can be reconstructed from its multiple projections $\{\V_1,...,\V_m\}$, we treat the fused feature of these views as the learned feature of $\M$. 
Hence, we assume the feature of $\M$ as $E_{\M}(\M)=F\left(E(\V_1),...,E(\V_m)\right)$ and that of the transformed object $\widetilde{\M}$ as $E_{\M}(\widetilde{\M})=F\left(E(\widetilde{\V}_1),...,E(\widetilde{\V}_m)\right)$. 
Then, \eqref{eq:decode_fusion} can be interpreted as
\begin{equation}
  \hat{\t}=D\left[E_{\M}(\M), E_{\M}(\widetilde{\M})\right],
  \label{eq:inter_3d_ter}
\end{equation}
which essentially infers the 3D transformation from the feature representations of the 3D object before and after transformation.

Similar to the Fusion Scheme, we assume the feature of $\M$ as $E_{\M}(\M)=E(\V_i)$ and that of the transformed counterpart $\widetilde{\M}$ as $E_{\M}(\widetilde{\M})=E(\widetilde{\V}_i)$ in the Average Scheme.
That is, we push function $E(\cdot)$ to learn a good 3D representation from a single view as much as possible.
Thus, \eqref{eq:t_i} can also be interpreted by \eqref{eq:inter_3d_ter}.

We view the interpretation of the decoding schemes by \eqref{eq:inter_3d_ter} as implicit learning of 3D transformation equivariant representations, \ie,
\begin{equation}
  E_{\M}(\widetilde{\M})=E_{\M}(\t(\M))=\rho_{\M}(\t)E_{\M}(\M),
\end{equation}
where $\rho_{\M}(\t)$ is a homomorphism of 3D transformation $\t$ in the representation space.

\section{MV-TER: The Algorithm}
\label{sec:method}


Given a 3D object $\M$, we randomly draw a transformation $\t \sim \mathcal{T}$ and apply it to $\M$ to obtain a transformed $\widetilde{\M}$.
Then we have $m$ views $\mathcal{V}=\{\V_1,...,\V_m\}$ by projecting $\M$ to 2D views.
Accordingly, the views after the 3D transformation are $\widetilde{\mathcal{V}}=\{\widetilde{\V}_1,...,\widetilde{\V}_m\}$.

To learn the applied 3D transformation $\t$, we design an end-to-end architecture as illustrated in Figure~\ref{fig:framework_fusion} for the \textbf{fusion} decoding scheme.
We choose existing CNN models as the representation learning module $E(\cdot)$ (\eg, AlexNet \cite{krizhevsky2012imagenet}, GoogLeNet \cite{szegedy2015going}), which extract the representation of each view separately.
The learned feature representations will be fed into a fusion module and a transformation decoding module $D(\cdot)$ respectively. 
The fusion module is to fuse the features of multiple views as the overall 3D object representation, \eg, by a view-wise max-pooling layer \cite{su2015multi} or group pooling layer \cite{feng2018gvcnn}.
The fused feature will serve as the general descriptor of the 3D object for the subsequent downstream learning tasks (\eg, classification and retrieval). 
The transformation decoding module $D(\cdot)$ is to estimate the 3D transformation parameters from the feature representations of multiple views. 
Next, we will discuss the representation learning module and transformation decoding module in detail.

\subsection{Representation Learning Module}

The representation learning module $E(\cdot)$ takes the original 2D views $\mathcal{V}$ and their transformed counterparts $\widetilde{\mathcal{V}}$ as the input.
$E(\cdot)$ learns feature representations of $\mathcal{V}$ and $\widetilde{\mathcal{V}}$ through a Siamese encoder network with shared weights.
Specifically, we employ the feature learning layers of a pre-trained CNN model as the backbone. 
Then, we obtain the features of each view before and after transformation. 

\subsection{Transformation Decoding Module}

To estimate the 3D transformation $\t$, we concatenate extracted features of multiple views before and after transformation at feature channel, which are then fed into the transformation decoder.
The decoder consists of several linear layers to aggregate the representations of multiple views for the prediction of the 3D transformation. 
As discussed in Section~\ref{subsec:decoding}, we have two strategies for decoding the transformation parameters. 
We can decode from the fused representations of multiple views before and after transformation as in \eqref{eq:decode_fusion}, or from each pair of original and equivariantly transformed views $\{\V_i, \widetilde{\V}_i\}$ to take average for final estimation as in \eqref{eq:final}.  
Based on the loss in \eqref{eq:loss}, $\t$ is decoded by minimizing the mean squared error (MSE) between the ground truth and estimated transformation parameters.

\section{Experiments}
\label{sec:experiments}

\begin{table*}[t]
  \centering
  \scriptsize
  \caption{3D object classification and retrieval results on ModelNet40 dataset.}
  \label{tab:results}
  \begin{tabular}{lccccc}
  \hline
  \multicolumn{1}{c}{\multirow{2}{*}{\textbf{Methods}}} &
    \multicolumn{2}{c}{\textbf{Training Configuration}} &
    \multirow{2}{*}{\textbf{Modality}} &
    \multirow{2}{*}{\begin{tabular}[c]{@{}c@{}}\textbf{Classification} \\ \textbf{(Accuracy)}\end{tabular}} &
    \multirow{2}{*}{\begin{tabular}[c]{@{}c@{}}\textbf{Retrieval} \\ \textbf{(mAP)} \end{tabular}} \\ \cline{2-3}
  \multicolumn{1}{c}{} & \textbf{Pre-train} & \textbf{Fine tune} &  &  &  \\ \hline
  VoxNet \cite{maturana2015voxnet} & - & ModelNet40 & voxels & 83.0 & - \\ 
  SubvolumeSup \cite{qi2016volumetric} & - & ModelNet40 & voxels & 89.2 & - \\ 
  Voxception-ResNet \cite{brock2016generative} & - & ModelNet40 & voxels & 91.3 & - \\ \hline
  PointNet \cite{qi2017pointnet} & - & ModelNet40 & points & 89.2 & - \\
  PointNet++ \cite{qi2017pointnet++} & - & ModelNet40 & points & 91.9 & - \\
  KD-Networks \cite{klokov2017escape} & - & ModelNet40 & points & 91.8 & - \\ \hline
  MVCNN \cite{su2015multi} & ImageNet1K & ModelNet40 & 12 views & 89.9 & 70.1 \\ 
  MVCNN, metric \cite{su2015multi} & ImageNet1K & ModelNet40 & 12 views & 89.5 & 80.2 \\
  MVCNN, multi-resolution \cite{qi2016volumetric} & ImageNet1K & ModelNet40 & 20 views & 93.8 & - \\
  RotationNet \cite{kanezaki2018rotationnet} & ImageNet1K & ModelNet40 & 12 views & 90.7 & - \\
  MHBN \cite{yu2018multi} & ImageNet1K & ModelNet40 & 12 views & 93.4 & - \\
  SeqViews2SeqLabels \cite{han2018seqviews2seqlabels} & ImageNet1K & ModelNet40 & 12 views & 93.4 & 89.1 \\
  3D2SeqViews \cite{han20193d2seqviews} & ImageNet1K & ModelNet40 & 12 views & 93.4 & 90.8 \\
  Relation Network \cite{yang2019learning} & ImageNet1K & ModelNet40 & 12 views & 94.3 & 86.7 \\
  MLVCNN, Center Loss \cite{jiang2019mlvcnn} & ImageNet1K & ModelNet40 & 36 views & 94.2 & 92.8 \\
  View-GCN \cite{wei2020viewgcn} & ImageNet1K & ModelNet40 & 12 views & 96.2 & - \\ \hline
  MVCNN (GoogLeNet) \cite{feng2018gvcnn} & ImageNet1K & ModelNet40 & 12 views & 92.2 & 74.1 \\
  MVCNN (GoogLeNet), metric \cite{feng2018gvcnn} & ImageNet1K & ModelNet40 & 12 views & 92.2 & 83.0 \\
  \textbf{MV-TER (MVCNN), average} & ImageNet1K & ModelNet40 & 12 views & \textbf{95.5} & \textbf{83.0} \\
  \textbf{MV-TER (MVCNN), fusion} & ImageNet1K & ModelNet40 & 12 views & \textbf{93.1} & \textbf{84.9} \\
  \textbf{MV-TER (MVCNN), average, Center Loss} & ImageNet1K & ModelNet40 & 12 views & \textbf{95.1} & \textbf{86.6} \\
  \textbf{MV-TER (MVCNN), fusion, Center Loss} & ImageNet1K & ModelNet40 & 12 views & \textbf{93.2} & \textbf{89.0} \\ 
  \cdashline{1-6}[2pt/2pt]
  GVCNN (GoogLeNet) \cite{feng2018gvcnn} & ImageNet1K & ModelNet40 & 12 views & 92.6 & 81.3 \\
  GVCNN (GoogLeNet), metric \cite{feng2018gvcnn} & ImageNet1K & ModelNet40 & 12 views & 92.6 & 85.7 \\
  \textbf{MV-TER (GVCNN), average} & ImageNet1K & ModelNet40 & 12 views & \textbf{97.0} & \textbf{88.8} \\
  \textbf{MV-TER (GVCNN), fusion} & ImageNet1K & ModelNet40 & 12 views & \textbf{96.4} & \textbf{88.3} \\
  \textbf{MV-TER (GVCNN), average, Center Loss} & ImageNet1K & ModelNet40 & 12 views & \textbf{95.7} & \textbf{91.5} \\
  \textbf{MV-TER (GVCNN), fusion, Center Loss} & ImageNet1K & ModelNet40 & 12 views & \textbf{96.3} & \textbf{91.1} \\
  \hline
  \end{tabular}
  \end{table*}
  
  In this section, we evaluate the proposed MV-TER model on two representative downstream tasks: 3D object classification and retrieval. 
  In particular, we apply the MV-TER to 3D object classification in Section~\ref{subsec:classification} and retrieval in Section~\ref{subsec:retrieval}. 
  Ablation studies are then conducted in Section~\ref{subsec:ablation} to demonstrate the effectiveness of the MV-TER under low labeling rates and few number of views.
  Further, we show the generalization performance on both synthetic and real-world datasets in Section~\ref{subsec:transfer}, and evaluate the 3D transformation estimation and feature maps in Section~\ref{subsec:further_eval}.
  
  
  \subsection{Datasets}
  
  We employ four datasets to evaluate the classification performance of the proposed MV-TER model, including two \textit{synthetic} datasets ModelNet40 \cite{wu20153d} and ShapeNetCore55 \cite{chang2015shapenet}, and two {\it real-world} multi-view datasets RGB-D \cite{lai2011large} and ETH-80 \cite{leibe2003analyzing}.
  We conduct 3D object classification and retrieval on the ModelNet40 dataset, and perform transfer learning on the ShapeNetCore55, RGB-D, and ETH-80 datasets, respectively.
  The details of these datasets are as follows.
  
  \textbf{ModelNet40} \cite{wu20153d}.
  This dataset contains $12,311$ CAD models from $40$ categories.
  We follow the standard training and testing split settings, \textit{i.e.}, $9,843$ models are used for training and $2,468$ models are for testing.
  
  \textbf{ShapeNetCore55} \cite{chang2015shapenet}. This dataset contains $51,162$ CAD models categorized into $55$ classes.
  The training, validation and test sets consist of $35,764$, $5,133$ and $10,265$ objects respectively.
  
  \textbf{RGB-D} \cite{lai2011large}. This dataset is a real-world multi-view dataset containing both RGB and depth images of $300$ objects from $51$ categories taken from multiple viewpoints.
  We sample $24$ views for each object instance for taining and testing.
  As suggested in \cite{lai2011large}, we perform $10$-fold cross validation to report the average accuracies.
  
  \textbf{ETH-80} \cite{leibe2003analyzing}. This dataset contains $80$ real-world models from eight categories. For each category, there are $10$ object instances and $41$ views for each object instance captured from different viewpoints.
  We employ the leave-one-object-out cross validation strategy as suggested in \cite{leibe2003analyzing} to report the classification accuracies.
  
  \textbf{Data preprocessing.}
  To acquire projected 2D views of the synthetic 3D object datasets ModelNet40 and ShapeNetCore55, we follow the experimental settings of MVCNN \cite{su2015multi} to render multiple views of each 3D object.
  Here, $12$ virtual cameras are employed to capture views with an interval angle of $30$ degree.
  Next, we employ rotation as our 3D transformations on objects and perform a random rotation with three parameters all in the range $[-180^{\circ}, 180^{\circ}]$ on the entire 3D object.
  We also render the views of the transformed 3D object using the same settings as the original 3D object. 
  The rendered multiple views before and after 3D transformations are taken as the input to our method.
  
  \begin{figure*}[t]
    \centering
    \includegraphics[width=0.8\textwidth]{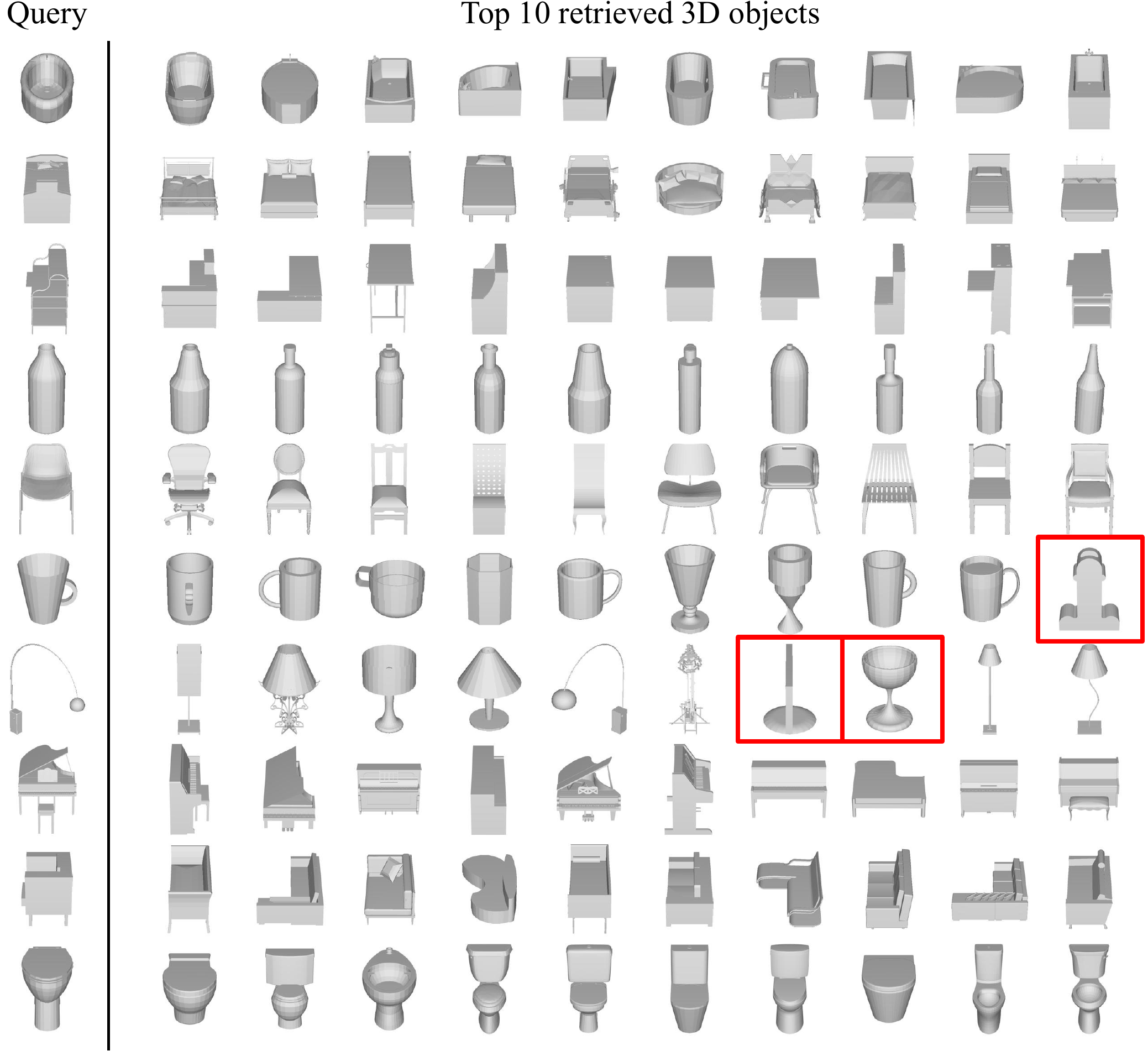}
    \caption{\textbf{3D object retrieval examples on ModelNet40 dataset.} Top $10$ matches are shown for each query, with mistakes highlighted in red.}
    \label{fig:retrieval}
  \end{figure*}
  
  \subsection{3D Object Classification}
  \label{subsec:classification}
  
  In this task, we employ the MV-TER as a self-supervisory regularizer to two competitive multi-view based 3D object classification methods: MVCNN \cite{su2015multi} and GVCNN \cite{feng2018gvcnn}, which are referred to as \textbf{MV-TER (MVCNN)} and \textbf{MV-TER (GVCNN)}. 
  Further, we implement with two transformation decoding schemes as discussed in Section~\ref{subsec:decoding}, including the \textbf{fusion} scheme and \textbf{average} scheme. 
  Then our model has four variants as presented in Table~\ref{tab:results}.
  
  \textit{Implementation Details}:
  We deploy GoogLeNet \cite{szegedy2015going} as our backbone CNN as in GVCNN \cite{feng2018gvcnn}.
  The backbone GoogLeNet is pre-trained on ImageNet1K dataset.
  We remove the last linear layer as the Siamese representation learning module to extract features for each view.
  Subsequent to the representation learning module, we employ one linear layer as the transformation decoding module.
  The output feature representations of the Siamese network first go through a channel-wise concatenation, which are then fed into the transformation decoding module to estimate the transformation parameters.
  The entire network is trained via the SGD optimizer with a batch size of $24$. 
  The momentum and weight decay rate are set to $0.9$ and $10^{-4}$, respectively.
  The initial learning rate is 0.001, which then decays by a factor of $0.5$ for every $10$ epochs.
  The weighting parameter $\lambda$ in (\ref{eq:loss}) is set to $1.0$.
  Also note that, MVCNN in Table~\ref{tab:results} has two variants with different backbones: MVCNN \cite{su2015multi} and MVCNN (GoogLeNet) \cite{feng2018gvcnn}. MVCNN \cite{su2015multi} uses the VGG-M \cite{chatfield2014return} as the backbone, while MVCNN (GoogLeNet) implemented in \cite{feng2018gvcnn} employs the GoogLeNet \cite{szegedy2015going}.
  In addition, RotationNet \cite{kanezaki2018rotationnet} and View-GCN \cite{wei2020viewgcn} are set up with $12$ views taken by the default camera system for fair comparison.
  
  \textit{Experimental Results}:
  As listed in Table~\ref{tab:results}, the \textbf{MV-TER (MVCNN), average} and \textbf{MV-TER (GVCNN), average} achieve classification accuracy of $95.5\%$ and $97.0\%$ respectively, which outperform the state-of-the-art View-GCN \cite{wei2020viewgcn}.
  Also, the \textbf{MV-TER (MVCNN), average} outperforms its baseline MVCNN (GoogLeNet) by $3.3\%$, while \textbf{MV-TER (GVCNN), average} outperforms the baseline GVCNN (GoogLeNet) by $4.4\%$, which demonstrates the effectiveness of our proposed MV-TER as a self-supervisory regularizer.

  \subsection{3D Object Retrieval}
  \label{subsec:retrieval}
  
  In this task, we directly employ the fused feature representations of \textbf{MV-TER (MVCNN)} and \textbf{MV-TER (GVCNN)} as the 3D object descriptor for retrieval.
  We denote $\mathbf{F}_{X}$ and $\mathbf{F}_{Y}$ as the 3D object descriptor of two 3D objects $\X$ and $\Y$ respectively, and use the Euclidean distance between them for retrieval.
  The distance metric is defined as
  \begin{equation}
      \text{dist}(\X,\Y)=\|\mathbf{F}_{X}-\mathbf{F}_{Y}\|_2.
  \end{equation}
  
  We take the mean average precision (mAP) on retrieval as the evaluation metric, and present the comparison results in the last column of Table~\ref{tab:results}.
  For MVCNN and GVCNN, a low-rank Mahalanobis metric learning \cite{su2015multi} is applied to boost the retrieval performance.
  In comparison, we train our MV-TER model without the low-rank Mahalanobis metric learning, but still achieve better retrieval performance, which validates the superiority of our feature representation learning for 3D objects.
  Further, we apply Triplet Center Loss \cite{he2018triplet} to our MV-TER.
  With Center Loss, our model further achieves an average gain of $3.3\%$ in mAP.
  As presented in the last column of Table \ref{tab:results}, the \textbf{MV-TER (GVCNN), average} and \textbf{MV-TER (GVCNN), fusion} achieve mAP of $91.5\%$ and $91.1\%$ respectively, which is comparable to MLVCNN with Center Loss \cite{jiang2019mlvcnn} while we only take $12$ views as input instead of $36$ views.
  We demonstrate some visual results of 3D object retrieval in Figure~\ref{fig:retrieval}.

  \subsection{Ablation Studies}
  \label{subsec:ablation}
  
  \subsubsection{On the Number of Views}
  
  We quantitatively evaluate the influence of number of views on the classification task.
  Specifically, we randomly choose $\{8,12\}$ views from all the views as the input to train \textbf{MV-TER (GVCNN), average} respectively, leading to two learned networks. 
  Then, we randomly select $\{2,4,8,12\}$ views from all the testing views to evaluate the classification accuracy of the two networks respectively, as reported in Table~\ref{tab:num_views}.
  We see that we constantly outperform GVCNN with different number of training views and testing views. 
  In particular, when the number of testing views reaches the extreme of two views for multi-view learning, our MV-TER model is still able to achieve the classification accuracy of $91.9\%$ and $91.2\%$, which outperforms GVCNN by a large margin.
  
  It is worth noting that though views are projected along fixed viewpoints in the training stage, the proposed model learns the transformation equivariant representations that capture the intrinsic and generalizable representations of multiple views, which is {\it independent of how the views are sampled}. This is validated by the experimental results in Table~\ref{tab:num_views}: we train the MV-TER on $8$ views and test on $12$ views where some views are not present in the training data. Results show that the MV-TER achieves the classification accuracy of $96.0\%$, which is comparable to training and testing on $12$ views.
  
  
  \begin{table}[t]
  \centering
  \scriptsize
  \caption{Experimental comparison between GVCNN and our \textbf{MV-TER (GVCNN), average}  with different number of input views for 3D object classification on ModelNet40 dataset. Here we use MV-TER for brevity.}
  \label{tab:num_views}
  \begin{tabular}{c|c|cc}
  \hline
  \multirow{2}{*}{\textbf{\begin{tabular}[c]{@{}c@{}}Training\\ Views\end{tabular}}} & \multirow{2}{*}{\textbf{\begin{tabular}[c]{@{}c@{}}Testing\\ Views\end{tabular}}} & \multicolumn{2}{c}{\textbf{Accuracy (\%)}} \\ \cline{3-4} 
   &  & \multicolumn{1}{c|}{\textbf{GVCNN}} & \textbf{MV-TER} \\ \hline
  \multirow{4}{*}{8} & 2 & \multicolumn{1}{c|}{71.2} & \textbf{91.9} \\
   & 4 & \multicolumn{1}{c|}{91.1} & \textbf{94.6} \\
   & 8 & \multicolumn{1}{c|}{93.1} & \textbf{95.4} \\
   & 12 & \multicolumn{1}{c|}{91.5} & \textbf{96.0} \\ \hline
  \multirow{4}{*}{12} & 2 & \multicolumn{1}{c|}{76.8} & \textbf{84.3} \\
   & 4 & \multicolumn{1}{c|}{90.3} & \textbf{92.5} \\
   & 8 & \multicolumn{1}{c|}{92.1} & \textbf{95.8} \\
   & 12 & \multicolumn{1}{c|}{92.6} & \textbf{97.0} \\ \hline
  \end{tabular}
  \end{table}

  \subsubsection{On Different Labeling Rates}
  
  We adopt six different label rates in the set $\{0.01,0.02,0.03,0.04,0.05,0.10\}$ to train four models for comparison: MVCNN (AlexNet), GVCNN (AlexNet), \textbf{MV-TER (MVCNN), average} and \textbf{MV-TER (GVCNN), average}.
  When training MVCNN (AlexNet) and GVCNN (AlexNet), we only use a small amount of labeled data to minimize the cross entropy loss for training, and then employ all the test data for evaluation. 
  When training \textbf{MV-TER (MVCNN), average} and \textbf{MV-TER (GVCNN), average}, we adopt all the data (labeled and unlabeled) to predict the 3D transformations without the use of labels, and then adopt only labeled data to acquire classification scores. 
  That is, we minimize \eqref{eq:loss} with a small amount of labels taken for the classification loss $\ell_{\text{task}}$. 
  In all the four models, a pre-trained AlexNet \cite{krizhevsky2012imagenet} on ImageNet1K dataset is employed as the backbone CNN.
  
  \begin{figure}[t]
    \includegraphics[width=0.95\columnwidth]{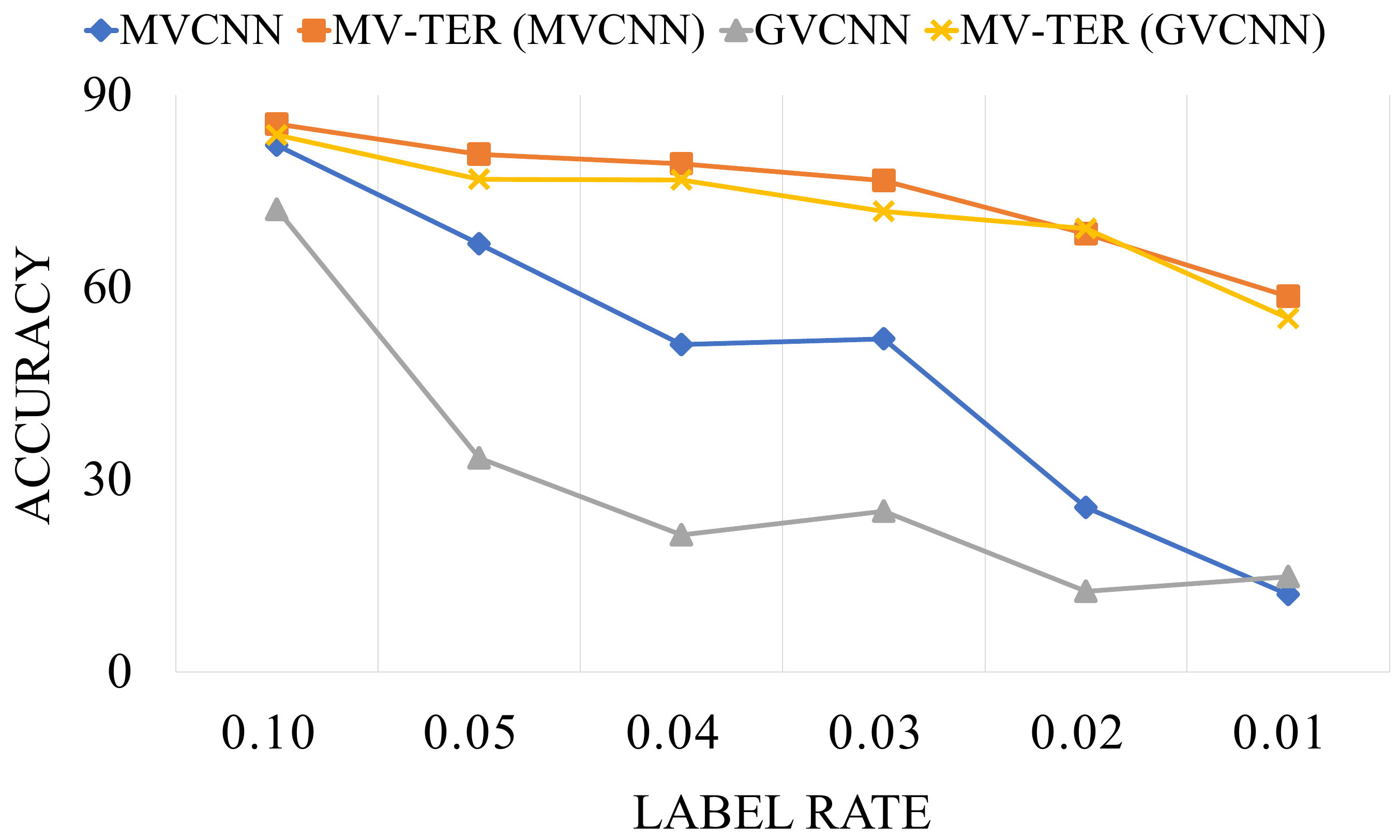}
    \caption{\textbf{Classification accuracy with different label rates.}}
    \label{fig:label_rate}
  \end{figure}
  
  Figure~\ref{fig:label_rate} presents the classiﬁcation accuracy under the six label rates on ModelNet40 dataset.
  When the label rate is $0.10$, we see that the four models achieve comparable results, which benefits from the pre-training of the backbone AlexNet.
  When the label rate keeps decreasing, the performance of both MVCNN and GVCNN drop quickly, while the MV-TER models are much more robust. 
  Even at the extremely low label rate $0.01$, \textbf{MV-TER (MVCNN), average} and \textbf{MV-TER (GVCNN), average} achieve the classiﬁcation accuracy of $58.6\%$ and $55.2\%$ respectively, thus demonstrating the robustness of the proposed MV-TER model. 
  

  \subsection{Transfer Learning}
  \label{subsec:transfer}
  
  We further show the generalization performance of the proposed MV-TER under \textbf{average} and \textbf{fusion} schemes.
  We take the same network architecture and parameter settings as in Section \ref{subsec:classification}, except that we set $\lambda=0.5$ in (\ref{eq:loss}).
  In particular, we train the MV-TER model on the ShapeNetCore55 dataset, and test on other datasets by a linear SVM classifier using the feature representations of dimension $2048$ obtained from the second last fully-connected layer of MV-TER.
  Table~\ref{tab:transfer} reports the classification comparison of MV-TER and two baseline methods on three datasets under the ShapeNetCore55 pre-training strategies.
  As we can see, the proposed MV-TER with two decoding schemes improves the average classification accuracy by $2.95\%$ and $2.85\%$ respectively on the synthetic dataset ModelNet40 under the ShapeNetCore55 pre-training strategy compared with the two baseline methods MVCNN and GVCNN, thus validating the generalizability.
  
  \begin{table}[t]
  \centering
  \scriptsize
  \caption{Classification comparison of MV-TER and two baseline methods under the ShapeNetCore55 pre-training strategy.}
  \label{tab:transfer}
  \begin{tabular}{cc|cc}
  \hline
  \textbf{Method} & \textbf{Accuracy (\%)} & \textbf{Method} & \textbf{Accuracy (\%)} \\ \hline
  \multicolumn{4}{c}{ModelNet40} \\ \hline
  MVCNN & $85.9$ & GVCNN & $87.9$ \\
  MV-TER, average & $\mathbf{88.7}$ & MV-TER, average & $\mathbf{91.2}$ \\
  MV-TER, fusion & $\mathbf{89.0}$ & MV-TER, fusion & $\mathbf{90.3}$ \\ \hline
  \multicolumn{4}{c}{RGB-D} \\ \hline
  MVCNN & $74.12 \pm 4.22$ & GVCNN & $76.41 \pm 4.08$ \\
  MV-TER, average & $\mathbf{75.75 \pm 2.59}$ & MV-TER, average & $\mathbf{78.56 \pm 3.52}$ \\
  MV-TER, fusion & $\mathbf{76.60 \pm 4.30}$ & MV-TER, fusion & $\mathbf{80.20 \pm 3.18}$ \\ \hline
  \multicolumn{4}{c}{ETH-80} \\ \hline
  MVCNN & $92.50 \pm 8.25$ & GVCNN & $93.75 \pm 10.08$ \\
  MV-TER, average & $\mathbf{93.75 \pm 6.25}$ & MV-TER, average & $\mathbf{97.50 \pm 5.00}$ \\
  MV-TER, fusion & $\mathbf{95.00 \pm 6.12}$ & MV-TER, fusion & $\mathbf{92.50 \pm 8.29}$ \\ \hline
  \end{tabular}
  \end{table}
  
  \begin{figure*}[htbp]
    \centering
    \includegraphics[width=0.65\textwidth]{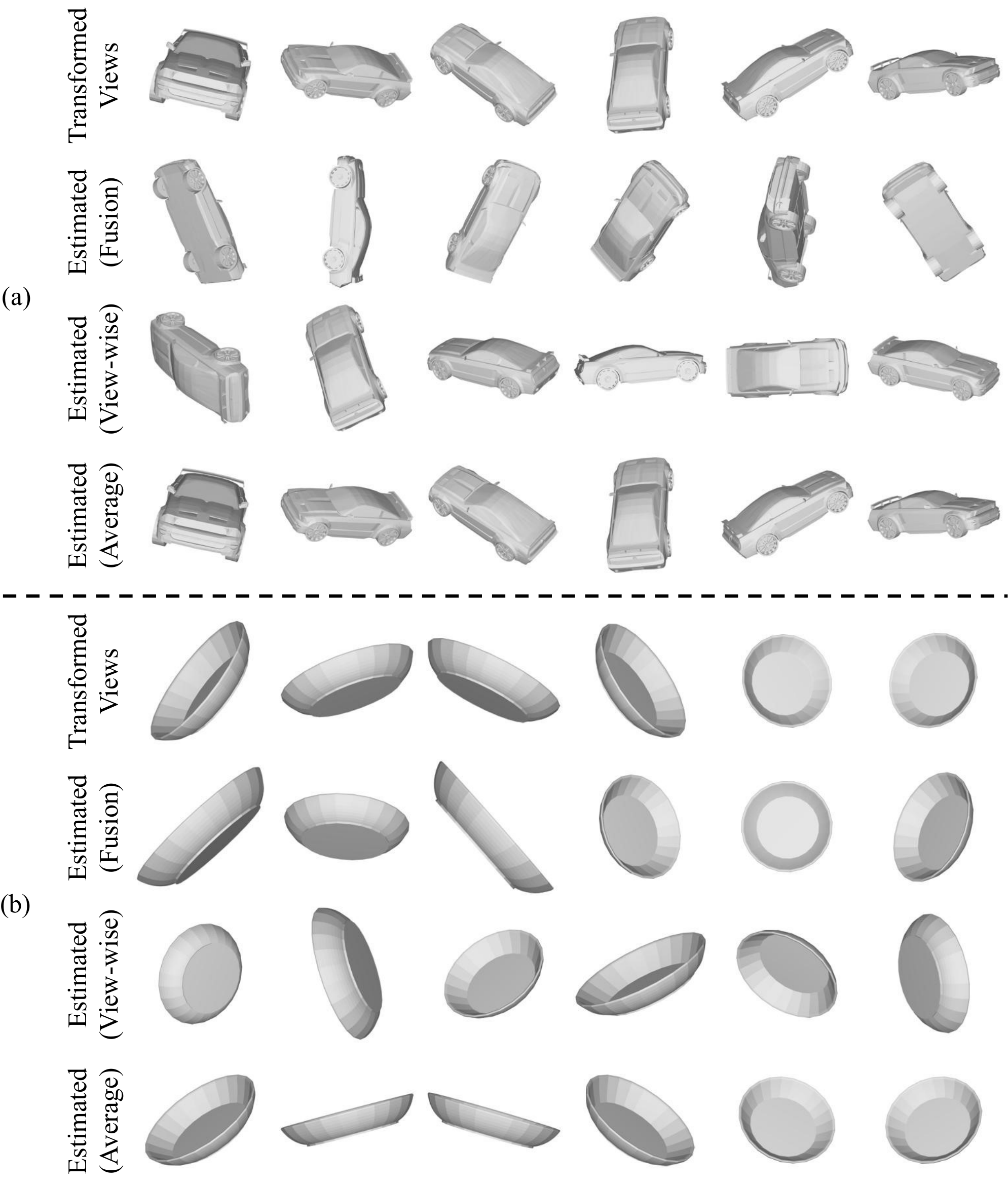}
    \caption{\textbf{Illustration of multiple views projected from 3D objects in the same posture: (a) Car and (b) Bowl.} The four rows of (a) and (b) demonstrate multiple views projected from the 3D object with the following 3D transformations applied respectively: 1) the ground-truth 3D transformation; 2) the estimated 3D transformation of the \textit{fusion} decoding scheme; 3) the individually estimated 3D transformations $\hat{\t}_i$'s from each view during the \textit{average} decoding scheme (with $\hat{\t}_i$ applied to the $i$th view); and 4) the finally averaged 3D transformation of the \textit{average} decoding scheme.}
    \label{fig:transformation}
  \end{figure*}
  
  We then evaluate our MV-TER on the {\it real-world} multi-view datasets RGB-D and ETH-80.
  As shown in the middle and bottom parts of Table~\ref{tab:transfer}, compared with the two baseline methods MVCNN and GVCNN, the MV-TER improves the average classification accuracy by $2.05\%$ and $2.97\%$ respectively on the RGB-D dataset, and $1.88\%$ and $1.25\%$ on the ETH-80 dataset. 
  Also, our classification results are more stable with less variation during the cross validation in general. 
  This demonstrates the generalizability of the MV-TER to real-world data.

  \subsection{Further Evaluation}
  \label{subsec:further_eval}
  
  \begin{figure}[t]
    \centering
    \includegraphics[width=0.75\columnwidth]{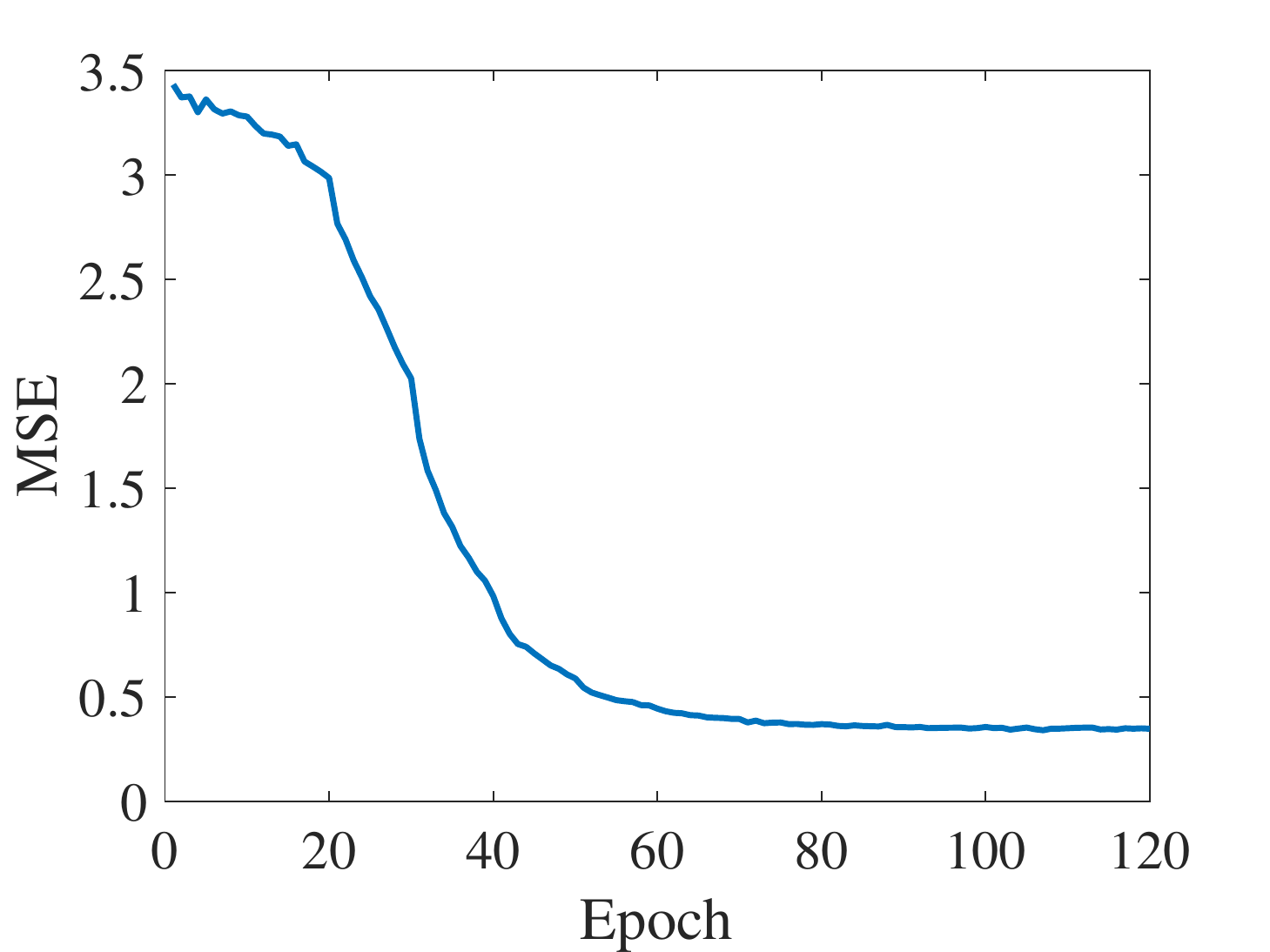}
    \caption{\textbf{Transformation estimation error from the \textit{average} scheme.}}
    \label{fig:mse_loss}
  \end{figure}
  
  \begin{figure*}[htbp]
      \centering
      \includegraphics[width=0.49\textwidth]{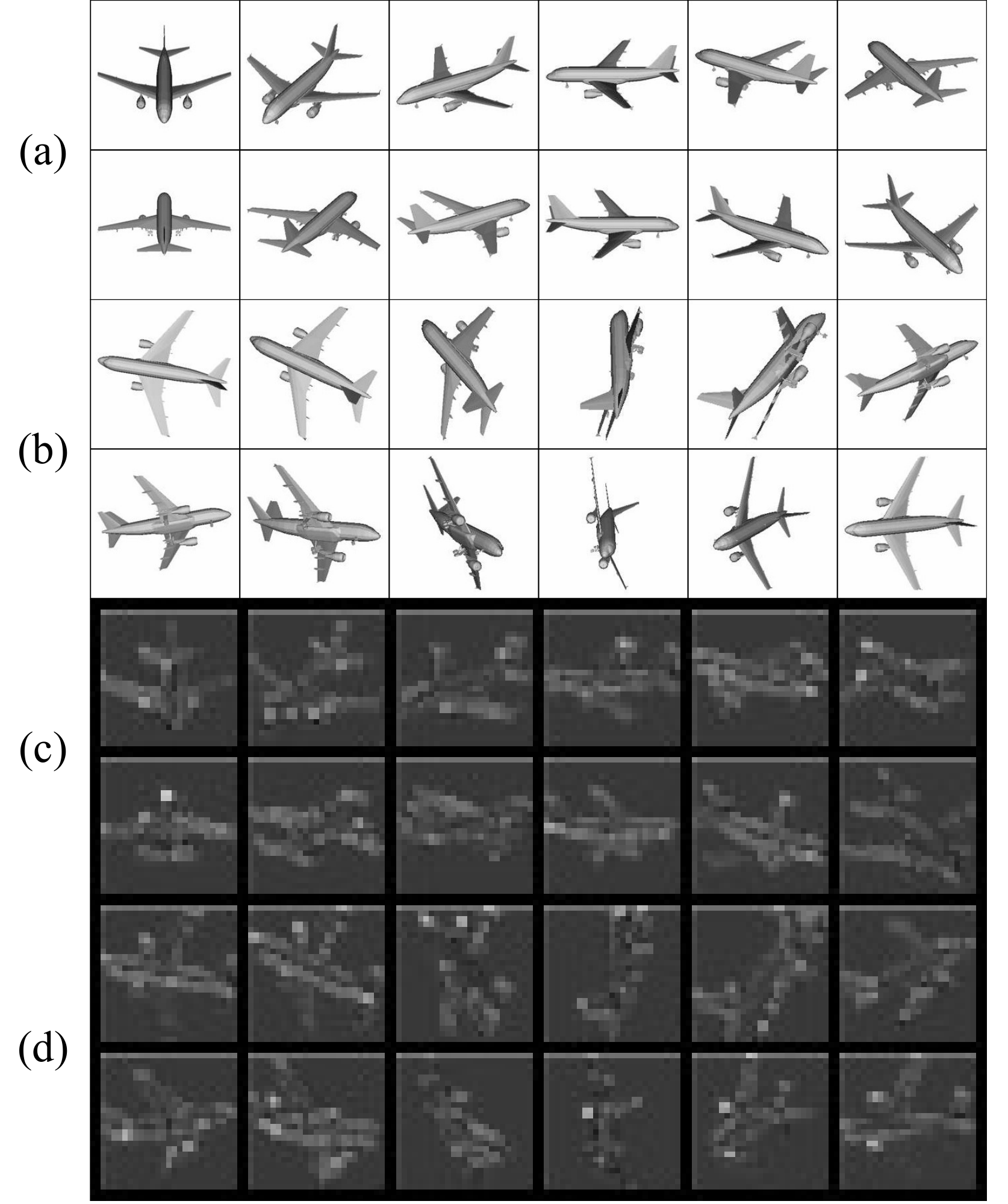}
      \includegraphics[width=0.49\textwidth]{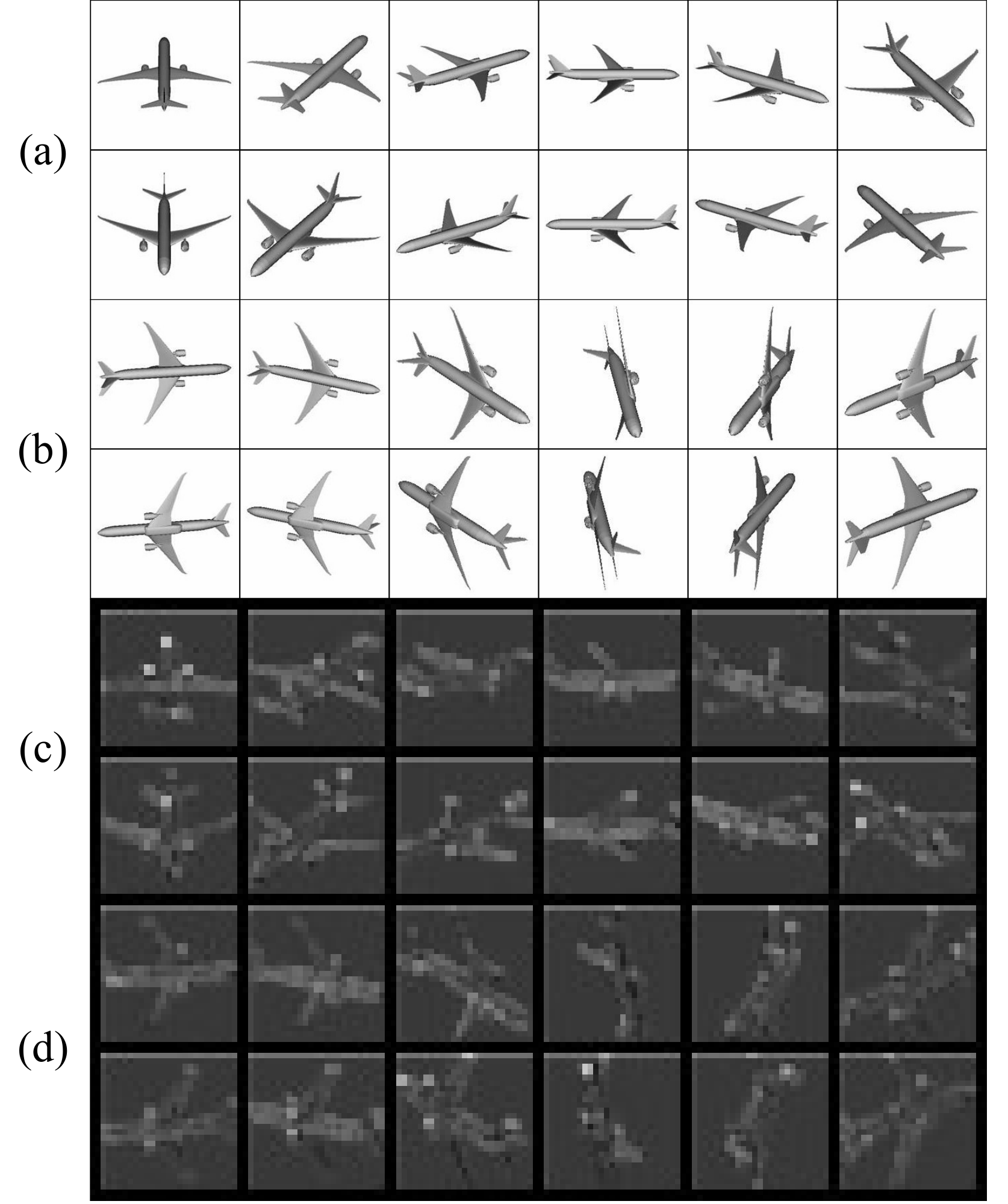}
      \caption{\textbf{Illustration of feature maps of multiple views projected from 3D objects before and after transformation in the \textit{same} category Airplane.} (a) and (b) demonstrate multiple views projected from the 3D object before and after transformations, respectively; (c) and (d) show the feature maps of the corresponding views above.}
      \label{fig:same_class_feats}
  \end{figure*}
  
  \begin{figure*}[htbp]
      \centering
      \includegraphics[width=0.49\textwidth]{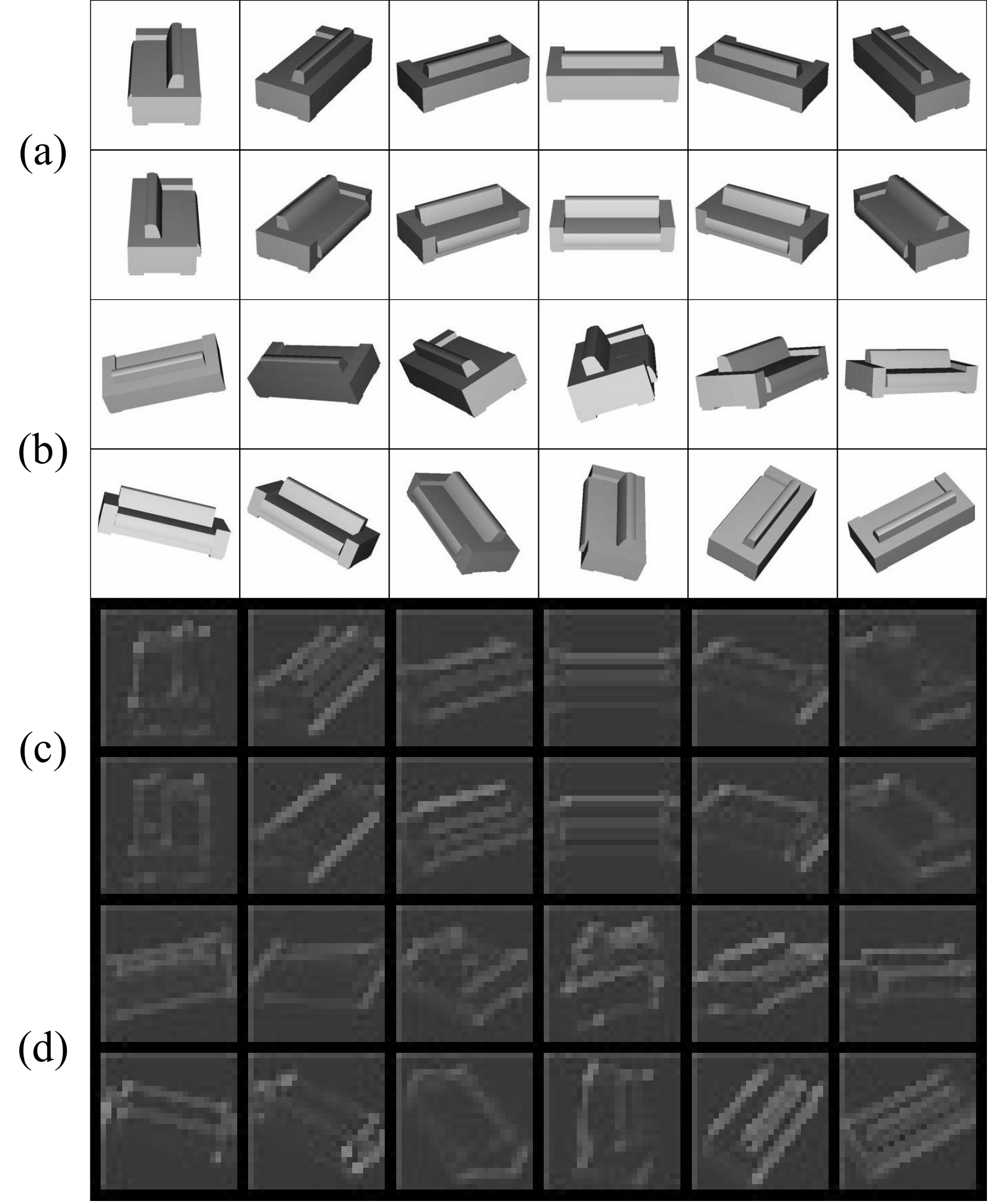}
      \includegraphics[width=0.49\textwidth]{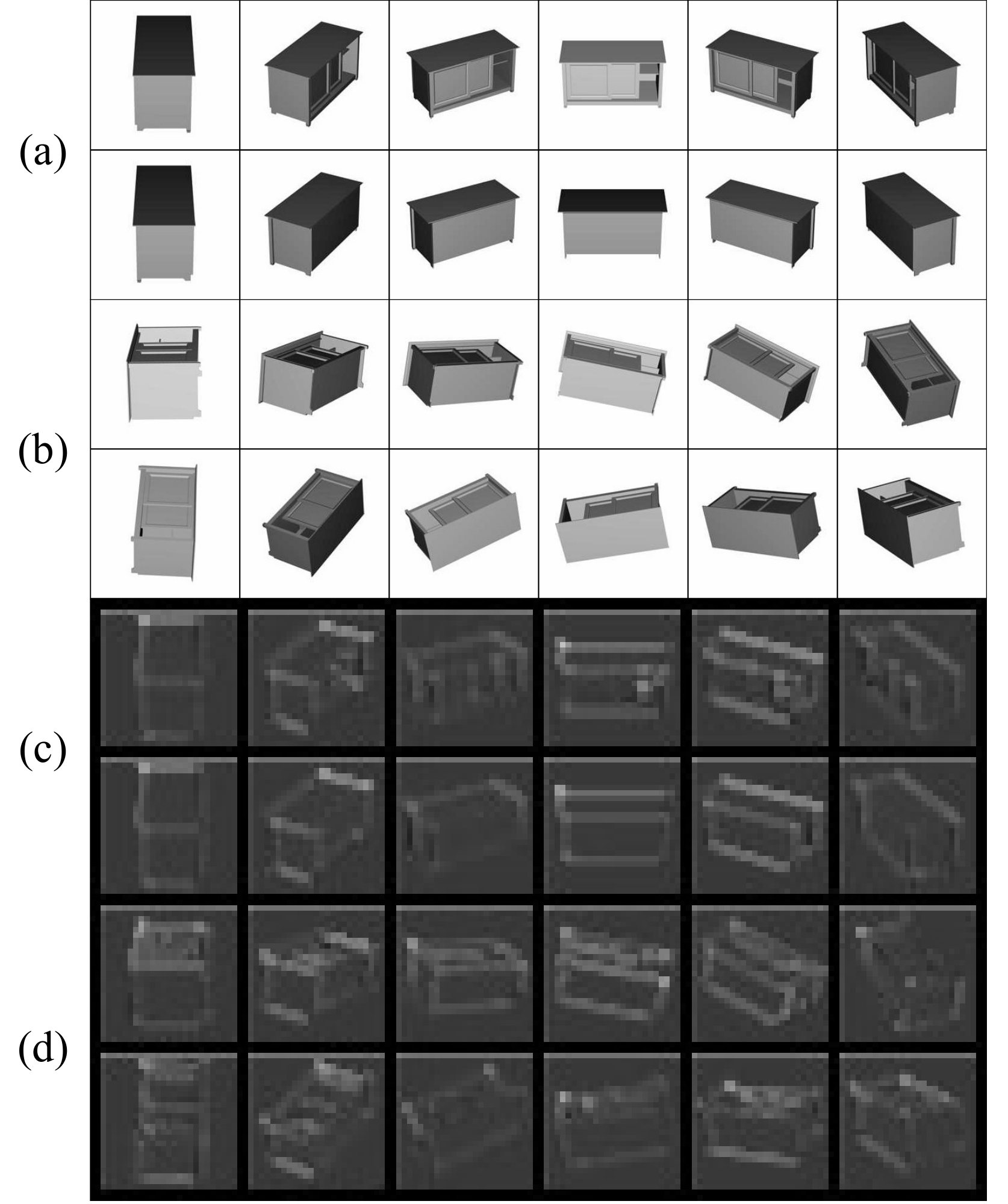}
      \caption{\textbf{Illustration of feature maps of multiple views projected from 3D objects before and after transformation in \textit{different} categories Sofa (left) and TV Stand (right).} (a) and (b) demonstrate multiple views projected from the 3D object before and after transformations, respectively; (c) and (d) show the feature maps of the corresponding views above.}
      \label{fig:diff_calss_feats}
  \end{figure*}
  
  \subsubsection{Evaluation of the 3D Transformation Estimation}
  \label{subsubsec:vis_transformation}
  
  Further, to intuitively interpret the estimated 3D transformations from the proposed \textit{fusion} and \textit{average} decoding schemes, we visualize the multiple views projected from 3D objects \texttt{Car} and \texttt{Bowl} with the estimated 3D transformations applied. 
  In Figure~\ref{fig:transformation}(a) and Figure~\ref{fig:transformation}(b), the first, second and fourth rows demonstrate the projected views from the 3D object with the {\it same} 3D transformation: the ground truth, the estimation from the \textit{fusion} scheme and the estimation from the \textit{average} scheme. 
  In the third row, each view is the result of each individually estimated 3D transformation $\hat{\t}_i$ as in \eqref{eq:t_i}, \ie, view-wise transformations. Note that each column is rendered under the same viewpoint.
  We see that our MV-TER model estimates more accurate 3D transformations via the \textit{average} scheme, which is consistent with the objective results.
  
  Moreover, Figure~\ref{fig:mse_loss} shows the transformation estimation error on the ModelNet40 dataset under the \textit{average} scheme. The horizontal axis is the index of the training epoch, while the vertical axis refers to the mean squared error. We observe that the MV-TER loss decreases rapidly in the first $40$ epochs. Until the $60$th epoch, the mean squared error basically converges to a very small number, thus validating the effectiveness of our model in the transformation estimation.
  
  \subsubsection{Visualization of Feature Maps}
  
  We visualize the feature maps of multiple views projected from 3D objects before and after transformation in Figure~\ref{fig:same_class_feats} for the same category and Figure~\ref{fig:diff_calss_feats} for different categories. We see that the feature maps of projected multiple views transform equivariantly with those of the input views.
  In Figure~\ref{fig:same_class_feats}, the feature maps from the same category are similar.  
  In contrast, in Figure~\ref{fig:diff_calss_feats}, although the 3D objects from two different categories are similar, their feature maps are discriminative. 
  This shows the robustness and effectiveness of the learned descriptor.

\section{Conclusion}
\label{sec:conclusion}

In this paper, we propose a novel self-supervised learning paradigm of Multi-View Transformation Equivariant Representations (MV-TER) via auto-encoding 3D transformations, exploiting the equivariant transformations of a 3D object and its projected multiple views. 
We perform a 3D transformation on a 3D object, which leads to equivariant transformations in projected multiple views. 
By decoding the 3D transformation from the fused feature representations of multiple views before and after transformation, the MV-TER enforces the representation learning module to learn intrinsic 3D object representations.  
Experimental results demonstrate that the proposed MV-TER significantly outperforms the state-of-the-art view-based approaches in 3D object classification and retrieval tasks.

\bibliographystyle{IEEEtran}

\end{document}